\documentclass{article}

     \usepackage[nonatbib, final]{neurips_2020}
\usepackage[numbers]{natbib}
\usepackage[utf8]{inputenc} 
\usepackage[T1]{fontenc}    
\usepackage{hyperref}       
\usepackage{url}            
\usepackage{booktabs}       
\usepackage{amsfonts}       
\usepackage{nicefrac}       
\usepackage{microtype}      
\usepackage{graphicx}
\usepackage{subcaption}
\usepackage{caption} 
\usepackage{booktabs}
\title{Turn Signal Prediction: A Federated Learning Case Study}

\author{%
  Sonal Doomra \\
  Ford Motor Company\\
  \texttt{sdoomra@ford.com} \\
   \And
   Naman Kohli \\
   Ford Motor Company \\
   \texttt{nkohli2@ford.com} \\
   \AND
   Shounak Athavale \\
   Ford Motor Company \\
   \texttt{sathaval@ford.com} \\
}

\begin{document}

\maketitle

\begin{abstract}
  Driving etiquette takes a different flavor for each locality as drivers not only comply with rules/laws but also abide by local unspoken convention. When to have the turn signal (indicator) on/off is one such etiquette which does not have a definitive right or wrong answer.  Learning this behavior from the abundance of data generated from various sensor modalities integrated in the vehicle is a suitable candidate for deep learning. But what makes it a prime candidate for Federated Learning are privacy concerns and bandwidth limitations for any data aggregation. This paper presents a long short-term memory (LSTM) based Turn Signal Prediction (on or off) model using vehicle control area network (CAN) signal data. The model is trained using two approaches, one by centrally aggregating the data and the other in a federated manner. Centrally trained models and federated models are compared under similar hyperparameter settings. This research demonstrates the efficacy of federated learning, paving the way for in-vehicle learning of driving etiquette.
  
\end{abstract}

\section{Introduction}

Neural networks have been used in engine controls for over twenty years but recently deep neural networks (DNNs) have found varied applications both in the enterprise and transportation mobility services \citep{SHAYLER1997899, luckow2016}. Current work of automotive related deep learning work is focused on leveraging sensor data produced by vehicles \citep{agarwal2020ford}. Vehicle sensors data is streamed via control area network (CAN) \citep{can} at speeds ranging from 1Gbit/s to 40GBit/s. To aggregate this data in a data center is challenging due to limited bandwidth, high latency and data breach risks \citep{carcost}. Therefore, it has led to the need for machine learning methods that are not only efficient in a large-scale industrial deployment but are also privacy-sustaining.   

\begin{figure}[!h]
\centering
\begin{subfigure}{.32\textwidth}
  \centering
\includegraphics[width=.98\linewidth]{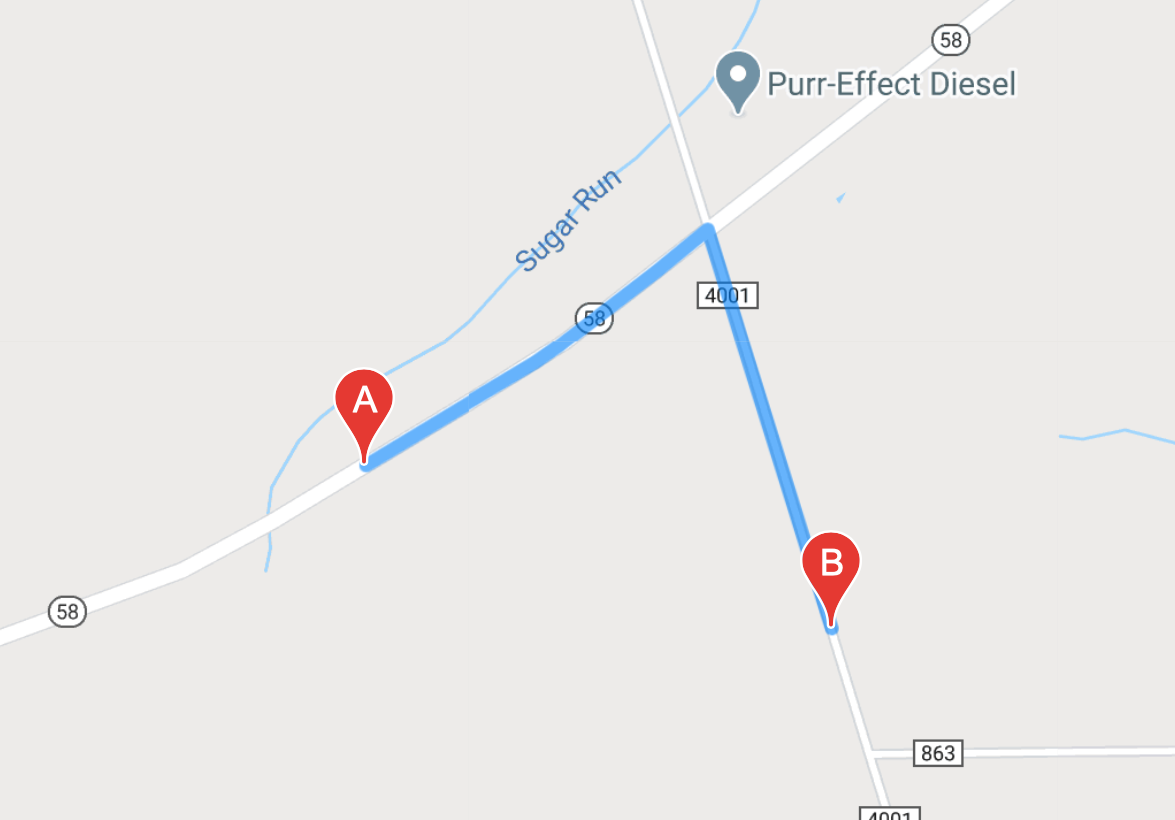}
\end{subfigure}%
\begin{subfigure}{.40\textwidth}
  \centering
\includegraphics[width=.98\linewidth]{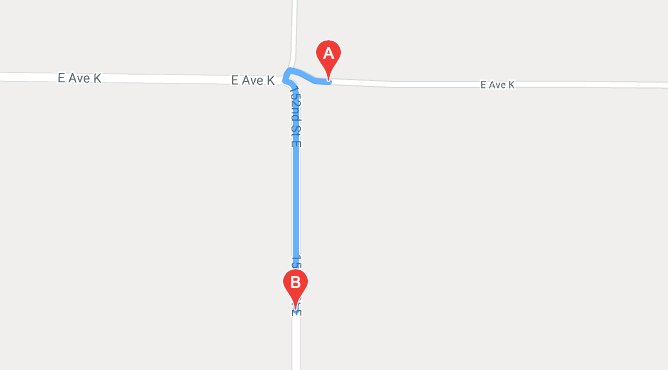}
\end{subfigure}%
\begin{subfigure}{.30\textwidth}
  \centering
\includegraphics[width=.98\linewidth]{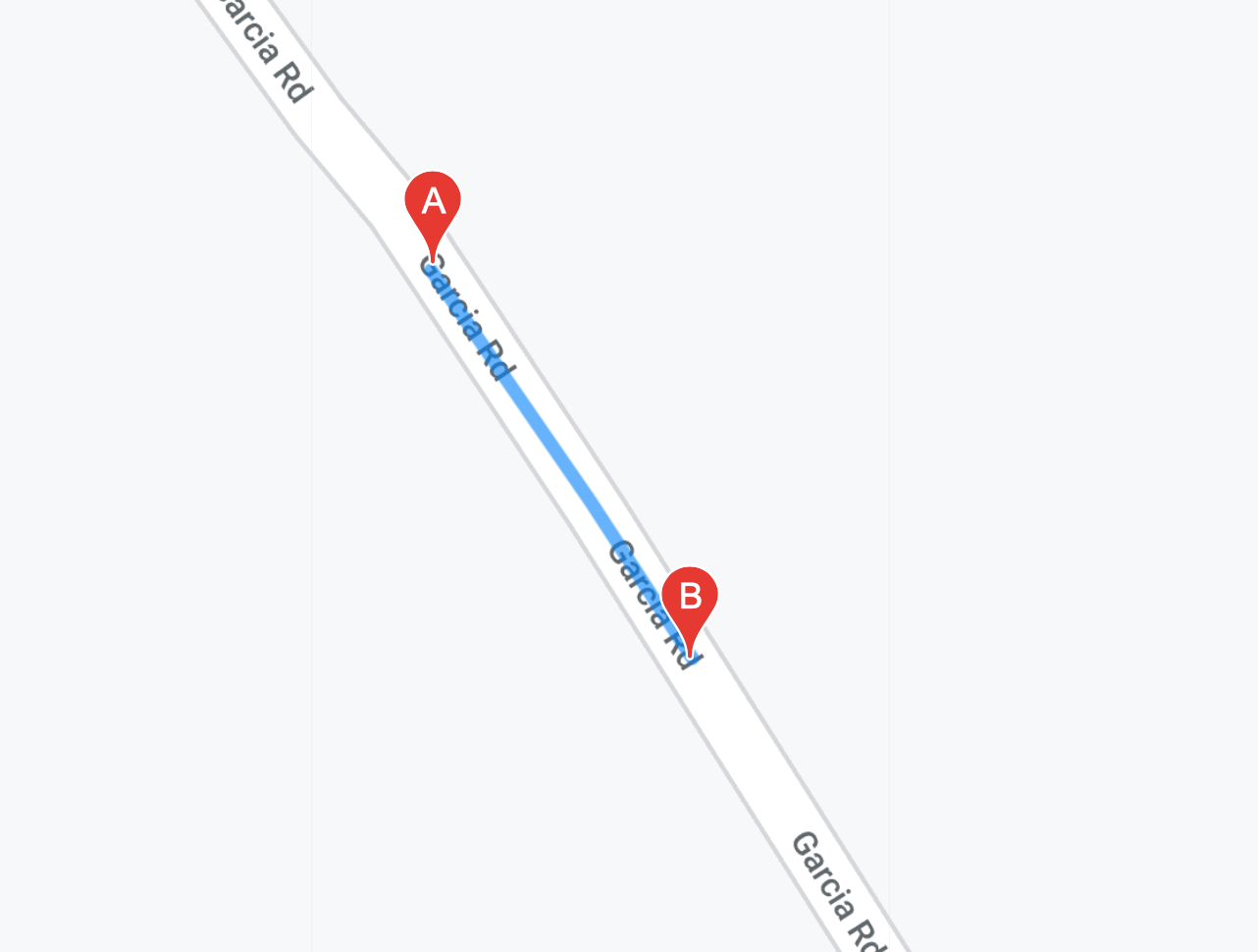}
\end{subfigure}%
\caption{Vehicle turns/lane changes (from A to B) prompting the off-on-off turn signal sequence}
\label{fig:1}
\end{figure}

Federated Learning \citep{Li_2020} is a variant of distributed machine learning algorithms where edge devices train individual AI models and share model parameters with the server instead of their data. Model parameters from edge devices are averaged to generate the global federated model that is redeployed back to the edge devices \citep{pmlr-v54-mcmahan17a}. This back and forth of model communications ensures privacy as data never leaves the device and drastically lowers the communication bandwidth. 

In this work, Federated Learning is utilized to simulate an “in-vehicle” use case for predicting driver behavior at the next time-step. Recommending drivers to switch on/off their turn indicator while driving is an application that would benefit by learning from a large number of drivers but at the same time faces the practical issues related to data transfer. Figure \ref{fig:1} displays some of the cases when turn signal indicator needs to be switched on when a vehicle is either making a turn (left or right) or changing lanes. The predictive model developed in this work can be used to alert the driver to turn on or off left/right indicator at an appropriate time. 

\section{Problem Description and Related Work}

\begin{figure}[!h]
\centering
\begin{subfigure}{.3\textwidth}
  \centering
\includegraphics[width=.98\linewidth]{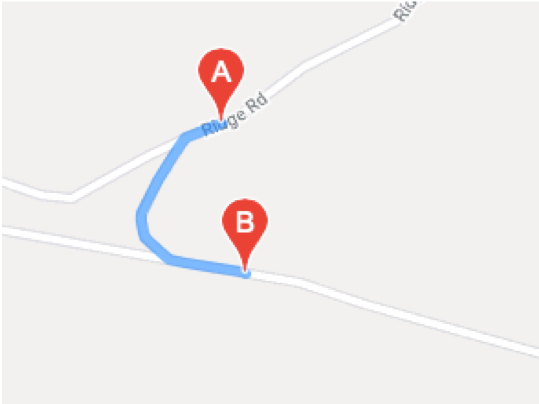}
\end{subfigure}%
\begin{subfigure}{.3\textwidth}
  \centering
\includegraphics[width=.98\linewidth]{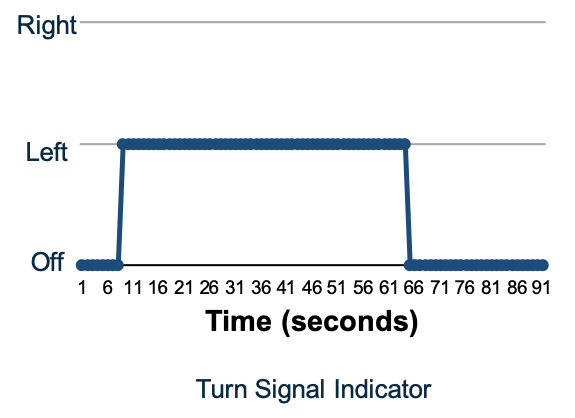}
\end{subfigure}%
\begin{subfigure}{.3\textwidth}
  \centering
\includegraphics[width=.98\linewidth]{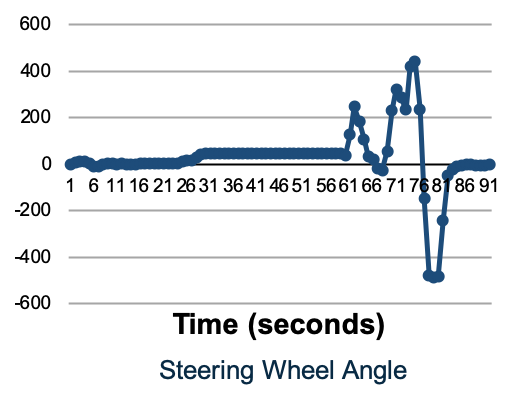}
\end{subfigure}
\begin{subfigure}{.3\textwidth}
  \centering
\includegraphics[width=.98\linewidth]{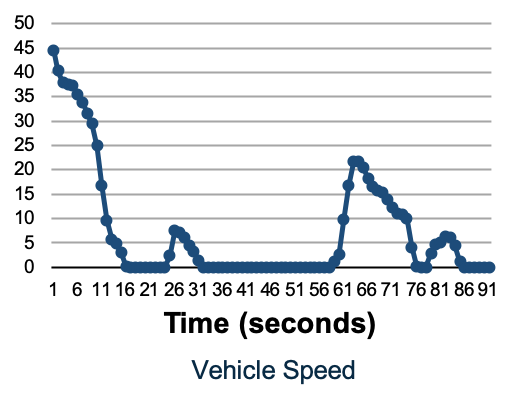}
\end{subfigure}%
\begin{subfigure}{.3\textwidth}
  \centering
\includegraphics[width=.98\linewidth]{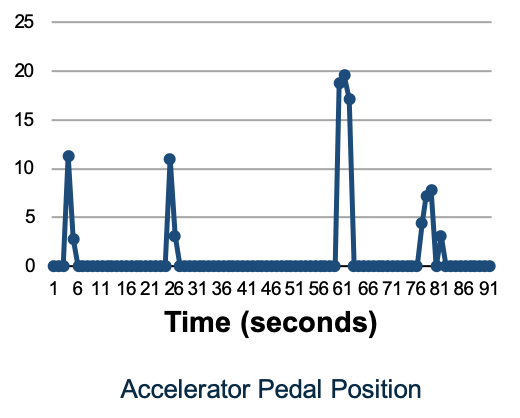}
\end{subfigure}%
\begin{subfigure}{.3\textwidth}
  \centering
\includegraphics[width=.98\linewidth]{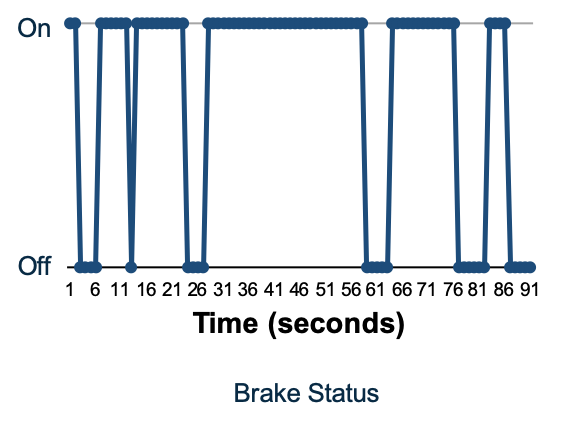}
\end{subfigure}%
\caption{As the vehicle is about to make a left turn (A-B), the driver indicates their intent of turning left by turning on the indicator. The plots depict change in signal value patterns as the vehicle makes the left turn.}
\label{fig:2}
\end{figure}

Turn signal indication does not have a precise answer to when it should be turned on or off.  Individual driving styles and forgetting to turn indicator on/off can catch other drivers off-guard.  A study by Society of Automotive Engineers published in 2012 \citep{Ponzani2012} showed that in the US alone, 48\% drivers do not use turn signals while changing lanes and 25\% while turning vehicles. Consequently, as much as 2 million car crashes are estimated to occur due to the neglected use of turn signals. Thus, a smart turn signal indicator in vehicles could help avoid them and enforce the right driving etiquette. Moreover, there are many different situations when the turn signal needs to be switched on, such as, when a driver is making a complete turn, changing lanes, or making a U turn. For instance, Figure \ref{fig:2} showcases the values of different CAN data signals as a vehicle is making a complete left turn. The regularities in CAN signals’ patterns can be learned to predict future turn indicator status.  

Since there is no right or wrong answer when it comes to turn signal timing, Federated Learning can help to learn a global behavior from drivers from different regions or countries while localizing it to adapt to local driving norms. Another important consideration for this task is that the data from vehicle's CAN bus like GPS coordinates is PII (Personally Identifiable Information) data and must be dealt with appropriately. Federated Learning gives an instant advantage of data privacy by ensuring that raw data never leaves the device (i.e. vehicles) it is stored on. 

Federated Learning was first introduced in \citep{KonecnyMRR16, Jakub2016, McMahanMRA16} by Google where they focused on building a distributed machine learning model that doesn’t leak data. Since then, a number of research papers have focused on improving the statistical challenges \citep{Smith2017, Zhao2018, yao2019federated} associated with training a global model, and improving the security aspect of the communication round \citep{Bonawitz2017, robin2017}. Federated Learning has been shown to converge in practical situations and has been used in multiple real-world applications like Google Swipe \citep{Hard2018}. There have also been efforts to utilize federated learning in personalization task \citep{Chen2018, jiang2019improving}. Federated learning has also been used in transfer learning where authors in \citep{Liu2020} utilized feature transfer learning for secure training. In this work, we evaluate if Federated Learning can be utilized for converging machine learning models for turn signal prediction. 

\section{Methodology}
The prediction task is designed as a classification task, where the turn signal status is determined at the next time step, given the multi-input signals collected at previous time steps. The data utilized in this research is a subset of data stored in Ford's Big Data Drive (BDD) dataset. This work utilizes tensorflow-federated \citep{TFF}, which is an open source framework for research and experimentation with Federated Learning. 

\subsection{Data Preprocessing}
This problem is formulated as a multi variate time series forecasting problem, where each signal from the CAN bus represents a time series. Multiple CAN signals were identified as useful features for predicting the turn signal status at a given time, such as Steering Wheel Angle (in degrees), Vehicle Speed (in kph), Turn Signal Status (serves as label for next time step), Accelerator Pedal Position, Brake Status, GPS coordinates, and Vehicle Direction (in degrees). Since the driver switches on the turn status only in specific situations, trips with turn signal status sequence off-on-off needs to be extracted for training \& test data sets. 

The raw signal data collected in the BDD are recorded at different sampling frequencies. This required a few pre-processing steps before transforming those signals into useful features for the deep learning model. To begin with, all signal frequencies were sampled to one second (or 1Hz), missing values for continuous variables were linearly interpolated and missing categorical values were set to the last known value. The pre-processing steps were performed for each vehicle individually. Only the trips with continuous timestamps were included in the final data set. Every trip consisted of 40 seconds of data before the turn signal was on (for look-back) and 10 seconds of data after it was turned off.  

\begin{figure}[!h]
\centering
\begin{subfigure}{.5\textwidth}
  \centering
\includegraphics[width=0.98\linewidth]{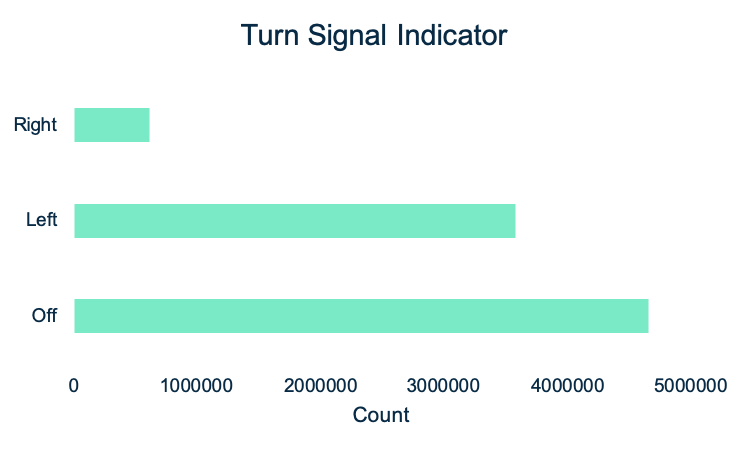}
\end{subfigure}%
\begin{subfigure}{.5\textwidth}
  \centering
\includegraphics[width=0.98\linewidth]{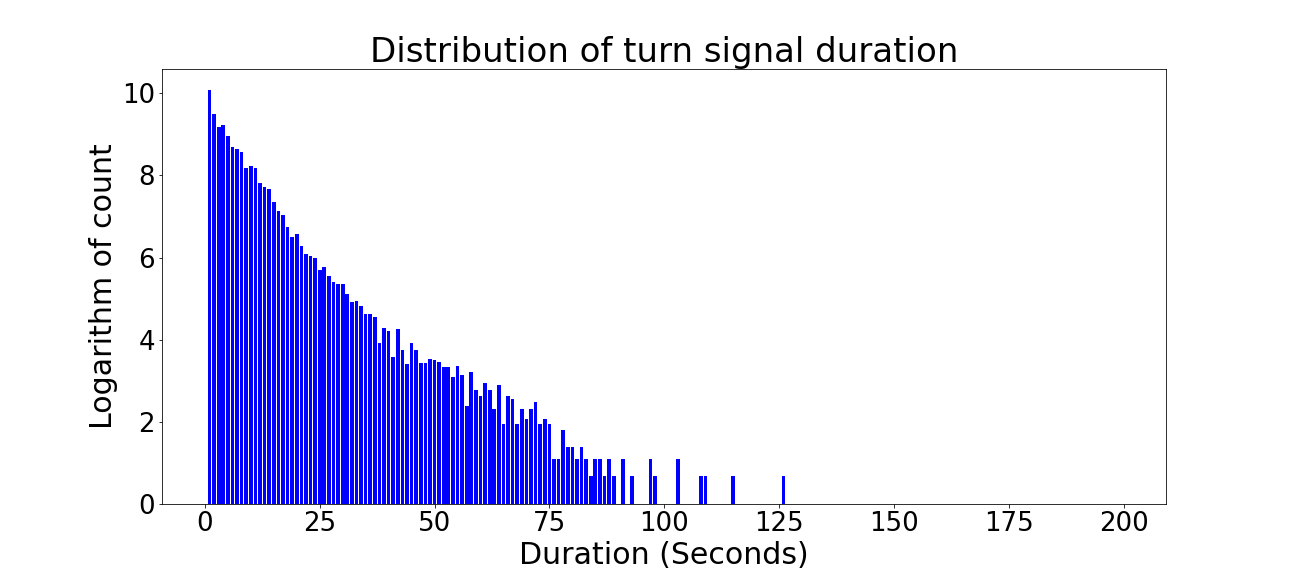}
\end{subfigure}
\caption{Distribution of Turn Signal Indicators data in training set}
\label{fig:3}
\end{figure}

With the above preprocessing steps, the dataset was successfully prepared with 66 distinct vehicles. Our final train dataset has about 8M training data points and 700K data points in the test set. Figure \ref{fig:3} shows data distribution for turn signal status in our final training dataset. It can be inferred from Figure \ref{fig:3} that the turn signal indicator is on for about the same number of times as it is off. However, since the BDD data lake consists of vehicles primarily from USA where right turn is mostly free, right indicator is on for a smaller number of times than the left indicator. This distribution is maintained in the test dataset as well. Figure \ref{fig:3} also shows a distribution of duration for how long the turn signal status is ON. Interestingly, majority of the turn signal duration is in the range of 0-100 seconds. There are certain cases when the duration is significantly less which could correspond to lane changes/quick turns and certain cases when the duration is significantly high which could indicate vehicle at traffic light or driver forgot to turn signal off.  

\begin{figure}[!t]
\centering
\includegraphics[width=0.60\linewidth]{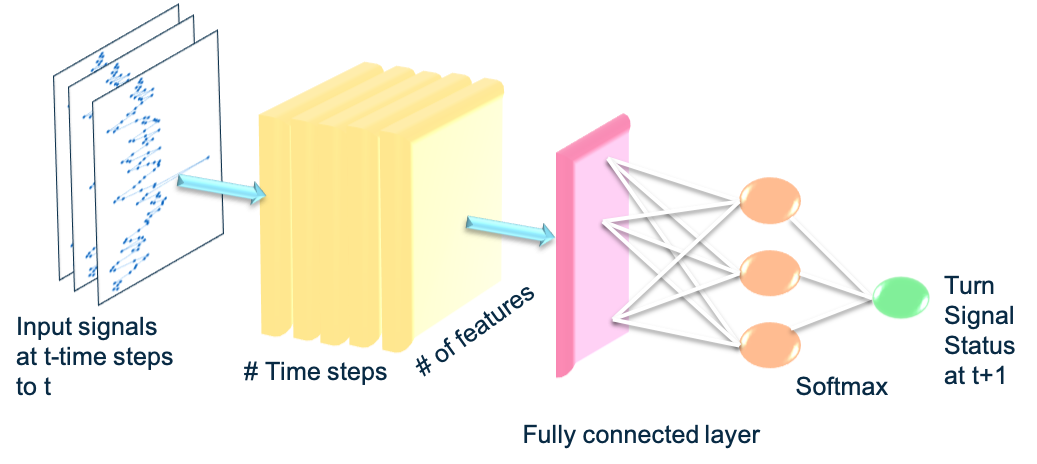}
\caption{LSTM model architecture for turn signal prediction task. The output layer of the fully connected network has three states, each corresponding to left, right or off state of the turn signal.}
\label{fig:4}
\end{figure}

\subsection{Model Architecture}
A multi-variate Long Short-term memory (LSTM) \citep{Schmidhuber1997} network is chosen as the model architecture for modeling the Turn Signal prediction problem because of the temporal nature of the data as shown in Figure \ref{fig:4}. The input to the LSTM network is a multi-step time series of multiple features (Steering wheel angle, vehicle speed, etc.). The output is a 3-dimensional vector, each of which corresponds to the state of turn signal indicator value at the future time step (t+1). The output layer is followed by a softmax activation function to get the final prediction.  

\subsection{Design of Experiments}
\begin{figure}[!h]
\centering
\includegraphics[width=.98\linewidth]{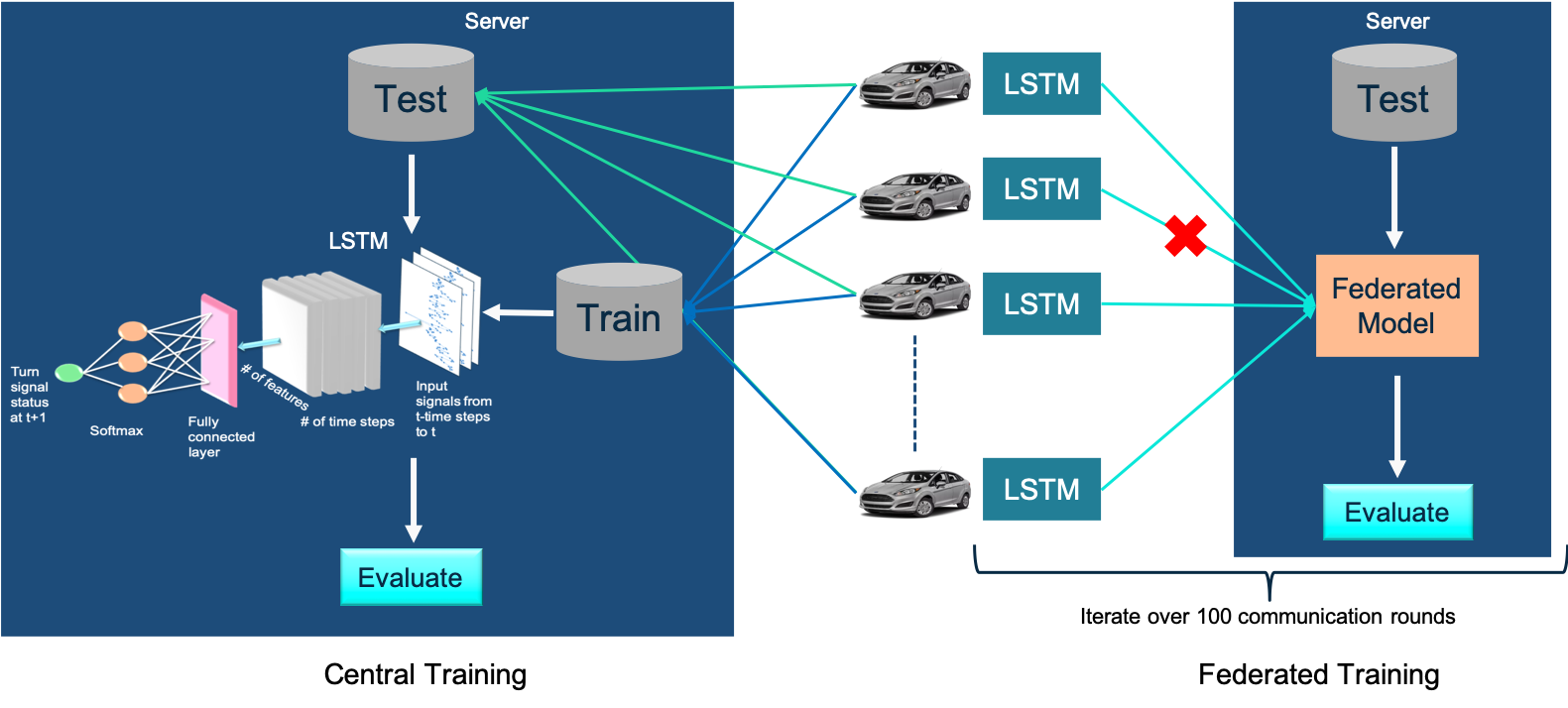}
\caption{For centralized training, data from each client is stored in the train and test set. For federated training, each vehicle is treated as a client and each local LSTM model is trained using local data. The client models are federated at the server to form a global model. Clients are randomly chosen for participation during the training step every round. (red x represents the vehicles not participating in that individual round, but eventually all will participate via multiple rounds)} 
\label{fig:5}
\end{figure}

To understand the performance of federated learning, the results from two sets of experiments were compared – one on centralized data training and the other on federated data training, using the design setup shown in Figure \ref{fig:5}. For centralized training, data from each vehicle is divided to form training, validation and test datasets such that there is no overlap in each of these datasets. For federated training, data from each vehicle is used to simulate a client. Aggregation at the server is performed using FedAvg algorithm as introduced in \citep{pmlr-v54-mcmahan17a}. F1-score and test accuracy are used as evaluation metrics for both set of experiments. The hyperparameters chosen for federated learning models are a subset of the ones used for centralized training as shown in Table \ref{table:hyper}. Adam is used as the optimizer in both sets of experiments.

\begin{table}[h]
\centering
\caption{Hyperparameters utilized for training LSTM model for turn signal prediction}
\begin{tabular}{@{}lcc@{}}
\toprule
                    & Central               & Federated              \\ \midrule
Batch Size          & 64, 128, 256          & 64, 128                \\
Window (time steps) & 5, 10, 15, 20, 30, 40 & 5                      \\
Hidden Units        & 50, 100, 150          & 50                     \\
Learning Rate       & $1\mathrm{e}{-3}$,$1\mathrm{e}{-4}$,$1\mathrm{e}{-5}$  & $1\mathrm{e}{-3}$, $1\mathrm{e}{-4}$, $1\mathrm{e}{-5}$ \\
Num Clients         & N/A                     & 10,25,all              \\
Num Rounds         & N/A                     & 100              \\
Local Epochs        & N/A                     & 1,5,10                 \\ \bottomrule
\end{tabular}
\label{table:hyper}
\end{table}


\section{Results and Analysis}
\begin{figure}[!h]
\centering
\begin{minipage}{.5\textwidth}
  \centering
  \includegraphics[width=.98\linewidth]{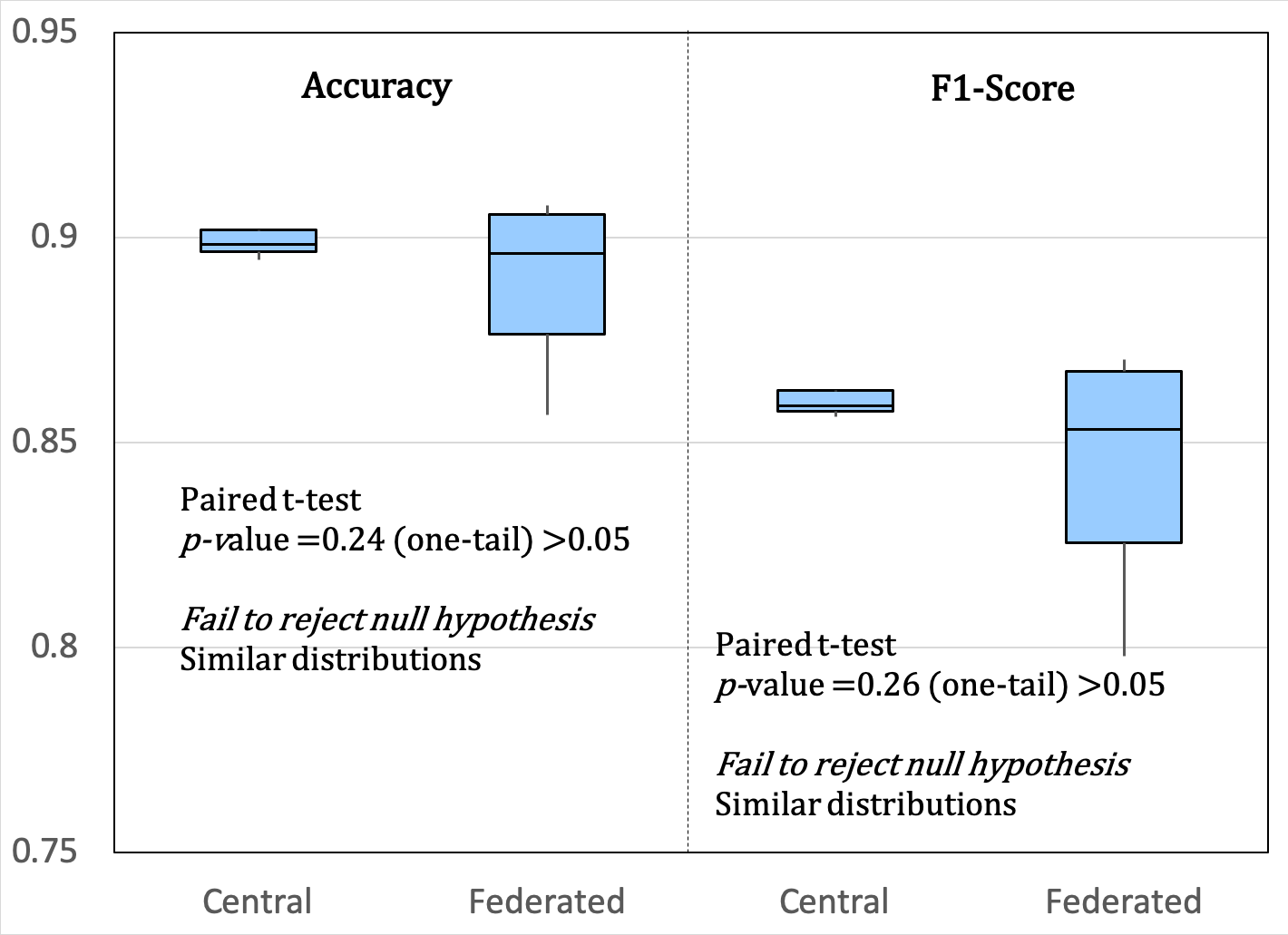} \\
  (a) Centralized training vs Federated training 
  \label{fig:test1}
\end{minipage}
\caption{Turn Signal Prediction performance across various hyperparameters for both centralized and federated learning}
\label{fig:6}
\end{figure}

This section presents empirical results for the experiments described in the previous section. For our purpose, an exhaustive grid search was performed on hyperparameters to find the optimal combination of model hyperparameters that results in most accurate performance on the test dataset. To evaluate our models’ performances, average F1-scores over all prediction classes (right, left and off) and test set accuracy were chosen as evaluation metrics. From the centralized training experiments, it was found that the window size of 5 seconds was best for our application. Figure \ref{fig:6} shows the distribution of test accuracy and average F1-scores of all models for both centralized and federated training experiments across various hyperparameters used in training. The paired t-test on test accuracy and average F1-scores for both central and federated experiments confirm that the distributions are similar. There is no significant difference in the accuracies and average F1-scores of central and federated models. In federated learning, there is more variation in the result distributions because the number of clients participating per round are also varied as a part of hyperparameter tuning. When small number of clients (for instance, only 10) participate every round, the performance of the model measured after 100 communication rounds is lower as compared to when all clients participate every round. This is because the data used for overall training is less.

For Federated Learning experiments, window size of 5 seconds is utilized for the client LSTM models as determined from the central training experiments. For Federated Learning experimentation, a subset of hyperparameters from central training experiments are chosen. One important hyperparameter to be considered for Federated Learning is the number of clients participating every round. It simulates the practical scenario where not all clients (vehicles) would be present at the same time for training and communicating their local updates to the server. When 25 or all clients participated in every round, similar high performance is observed for those models. However, when only 10 clients participate every round, the global model accuracy is lower for some combinations of hyperparameters. The best federated model showed similar performance as the best centrally trained model on the same held-out test dataset as shown in Table \ref{table:1}. It is observed that both the models performed better in predicting the 'Off' status which is also evident from the classwise F1-score shown in Table \ref{table:1}. The model accuracy can be improved by adding more turn signals (left and right) in the training dataset. 

{
\renewcommand{\arraystretch}{1.25}
\begin{table}[!t]
\centering
\caption{Test accuracy and F1 scores of best centrally trained and global federated model}
\begin{tabular}{c|cccc|cccc}
\hline
Approach       & \multicolumn{4}{c|}{Accuracy(\%)}                                                    & \multicolumn{4}{c}{F1-Score}     \\ \hline
 & \begin{tabular}[c]{@{}c@{}}Overall \\ Weighted\end{tabular} & Off & Left & Right & Average & Off & Left  & Right \\ \hline 
Central        & 90.4                                                        & 92.3    & 85.6 & 86.5  & 0.866   & 0.932   & 0.859 & 0.807 \\ 
Federated      & 91.1                                                        & 93.9    & 85.4 & 82.3  & 0.874   & 0.938   & 0.859 & 0.826 \\ \hline
\end{tabular}
\label{table:1}
\end{table}
}

\begin{figure}[!h]
\centering
\begin{subfigure}{.4\textwidth}
  \centering
\includegraphics[width=.95\linewidth]{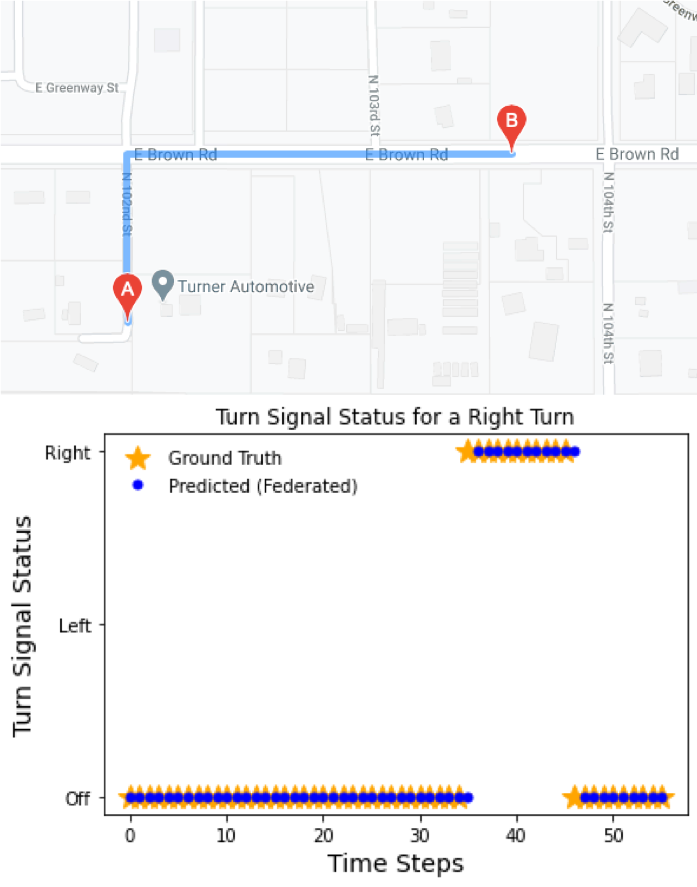}
\caption{}
\label{fig:7a}
\end{subfigure}%
\begin{subfigure}{.4\textwidth}
  \centering
\includegraphics[width=.95\linewidth]{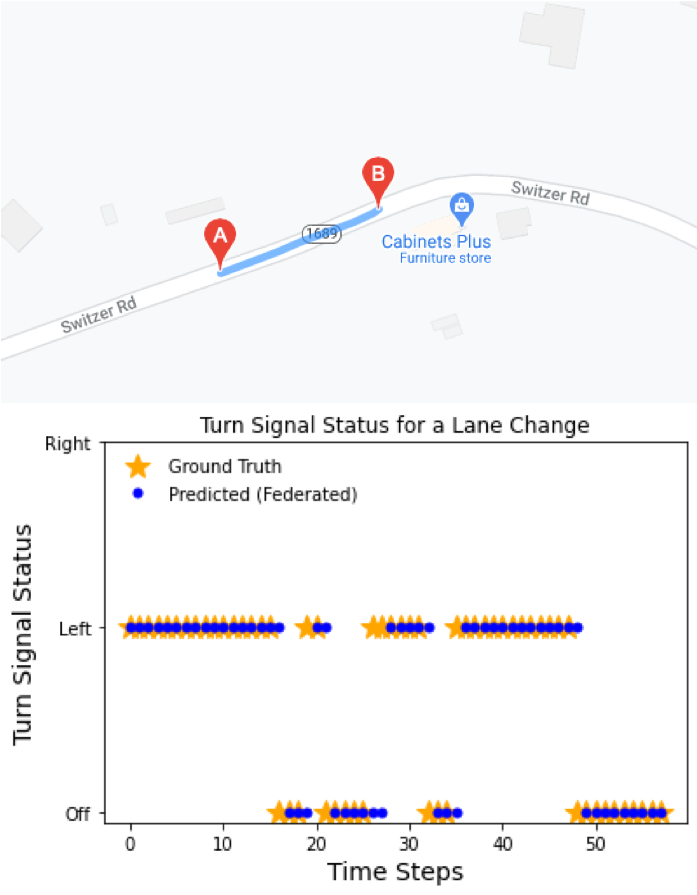}
\caption{}
\label{fig:7b}
\end{subfigure}
\caption{ a) A vehicle making a right turn and federated model predicting correctly. b) A vehicle makes a left lane change and the federated model predicts that left turn signal should be on. }
\label{fig:7}
\end{figure}

Upon visually analyzing some of these turns in the test set, it is observed that the federated model is able to predict the correct turn indicator status as shown in Figure \ref{fig:7a}. Interestingly, our model is able to pick on lane changes and correctly indicates that turn signal should be on when changing lanes as can be seen from Figure \ref{fig:7b}. From Figure \ref{fig:7}, it is observed that the model predictions vary from the ground truth mostly in the beginning and the end of the turn signal changes. It is important to understand that in this problem, the ground truth does not necessarily hold true because the drivers may start the turn signal earlier than they need or sometimes even forget to switch it off long after they have made the turn. For us, the goal was not to match the ground truth but to learn the turn signal change behavior collaboratively from different driver behaviors.  

\begin{figure}[!h]
\centering
\begin{subfigure}{.4\textwidth}
  \centering
\includegraphics[width=.95\linewidth]{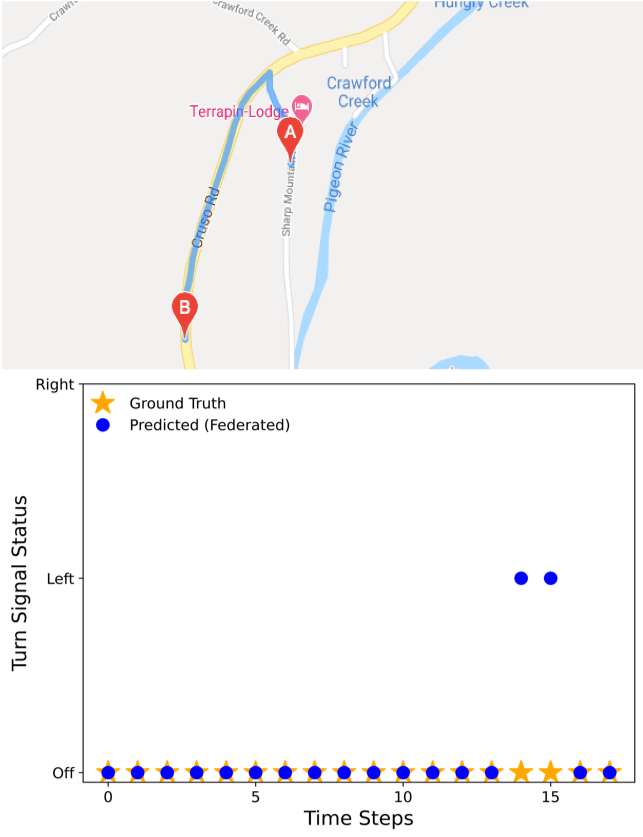}
\end{subfigure}%
\begin{subfigure}{.4\textwidth}
  \centering
\includegraphics[width=.95\linewidth]{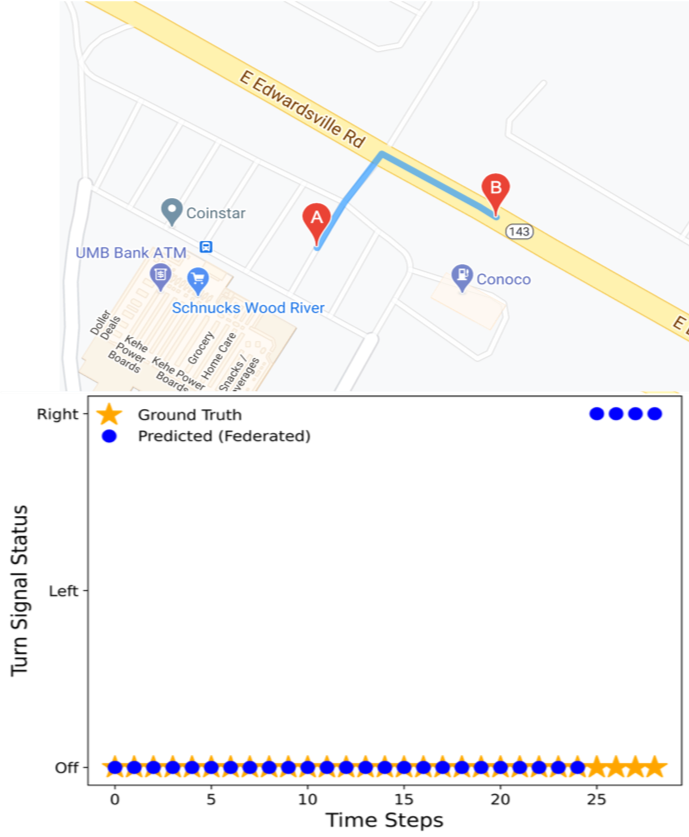}
\end{subfigure}
\caption{ a) A vehicle makes a left turn and the federated model predicts that left turn signal should be on when the driver did not switch on the turn signal. b) For a right turn the federated model predicts that the right turn signal should be on when the driver did not switch on the turn signal. }
\label{fig:8}
\end{figure}

Another interesting observation is that the federated model is able to predict that turn signal should be on even when the ground truth is off (i.e., the driver did not switch on the turn signal) and the vehicle is in fact making a turn. Figure \ref{fig:8} shows such scenarios with both a right turn and a left turn. Our federated model is able to predict the correct behavior in such cases because Federated Learning learns a shared model from multiple clients, driver patterns in our case, in a collaborative manner, thus learning a more accurate model from multiple different situations.

\section{Conclusions and Future Work}
In this research, we have trained a federated model to predict turn signal at next time step using vehicle CAN signal data from previous time steps. We observe that the performance of federated model is similar to a model that was trained centrally. This leads us to advantages of this solution wherein the data has never left the device (vehicle) preserving privacy and network bandwidth. Federated Learning, thus helps in utilizing machine learning in scenarios, especially learning in-vehicles, that were previously not considered due to privacy reasons or data bandwidth constraints.

The experiments presented in this paper were run in a simulated environment. The next phase of this research would be to move this into a real-world environment by setting up devices that stream the models from the vehicle. This would help us in evaluating practical challenges of Federated Learning such as communication frameworks, latency requirements, device scheduling and scheduling model training at the clients. Other aspects from the research side would be to look at the effect of heterogeneous clients where the data distribution of each client is varied and impact of adversaries during the federation.

\section{Broader Impact}
Virtual driver models that mimic local etiquette are essential for both virtual driving systems and driver assist features. This methodology can be used to develop these features. Further, transportation mobility services can be developed and tested. However, learning from multiple driving behaviors can create a biased system. Federated learning does help in mitigating bias in the global model by averaging local models but it must be tested broadly in different geographic locations. Additionally, having the turn signal on and off at appropriate time is a safety feature that can prevent accidents. Instead of monitoring driving behavior a posteriori using plug-in devices, the insurance companies can proactively assist the drivers. These models can clearly be extended using V2X data for others in the mobility ecosystem from pedestrians to various transportation modes. Although the Federated learning mitigates the communication and server storage data breach, individual data on the devices has to be protected \& purged. However, it does prevent mass scale data breaches that were observed recently.

\bibliographystyle{unsrtnat}
\bibliography{bibliography}
\end{document}